\documentclass{article}

\PassOptionsToPackage{usenames,dvipsnames}{xcolor}

\usepackage{amsmath, amssymb, amsthm}
\usepackage{mathtools}
\usepackage{bbm}
\usepackage{dsfont}

\usepackage{graphicx}

\usepackage{booktabs, array, multirow, makecell, tabularx}
\usepackage[usestackEOL]{stackengine}
\usepackage[colorlinks,linktoc=all,pagebackref=false]{hyperref}
\usepackage[all]{hypcap}
\hypersetup{citecolor=NavyBlue}
\hypersetup{linkcolor=NavyBlue}
\hypersetup{urlcolor=NavyBlue}
\usepackage[nameinlink,capitalise, noabbrev]{cleveref}
\creflabelformat{equation}{#2\textup{#1}#3}  % <- remove parenthesis from equations
\crefname{section}{\S}{\S\S}
\usepackage[acronym,nowarn,section,nogroupskip,nonumberlist]{glossaries}
\glsdisablehyper{}

\newacronym{dsm}{DSM}{Denoising Score Matching}
\newcommand{\bsx}{\boldsymbol{x}}

\newcommand{\calB}{{\mathcal{B}}}

\newcommand{\calD}{{\mathcal{D}}}

\newcommand{\calL}{{\mathcal{L}}}

\newcommand{\calN}{{\mathcal{N}}}

\newcommand{\bbE}{\mathbb{E}}

\theoremstyle{plain}% default

\theoremstyle{definition}

\newcommand{\KL}{D_\mathrm{KL}}

\newcommand{\given}{{\hspace{0.08em}|\hspace{0.08em}}}

\def\[#1\]{\begin{equation}\begin{aligned}#1\end{aligned}\end{equation}}

\newcommand{\umet}[1]{#1 ($\uparrow$)}
\newcommand{\dmet}[1]{#1 ($\downarrow$)}

\usepackage[accepted]{icml2024}

\usepackage{multirow}
\usepackage{algorithmic}
\usepackage{algorithm}

\icmltitlerunning{A Simple Early Exiting Framework for Accelerated Sampling in Diffusion Models}
\begin{document}

\twocolumn[
\icmltitle{A Simple Early Exiting Framework \\
for Accelerated Sampling in Diffusion Models}
% \icmltitle{Adaptive Score Estimation: Early Exiting for \\
%             Accelerated Inference in Diffusion Models}



\icmlsetsymbol{equal}{*}

\begin{icmlauthorlist}
\icmlauthor{Taehong Moon}{comp1}
\icmlauthor{Moonseok Choi}{sch}
\icmlauthor{EungGu Yun}{comp3}
\icmlauthor{Jongmin Yoon}{sch}
\icmlauthor{Gayoung Lee}{comp2}
\icmlauthor{Jaewoong Cho}{comp1}
\icmlauthor{Juho Lee}{sch,comp4}
\end{icmlauthorlist}

\icmlaffiliation{comp1}{KRAFTON}
\icmlaffiliation{comp2}{Naver AI Lab, South Korea}
\icmlaffiliation{comp3}{Independent researcher}
\icmlaffiliation{comp4}{AITRICS, South Korea}
\icmlaffiliation{sch}{Graduate School of AI, KAIST}

\icmlcorrespondingauthor{Juho Lee}{juholee@kaist.ac.kr}

\icmlkeywords{Machine Learning, ICML}

\vskip 0.3in
]



\printAffiliationsAndNotice{This work is partially done at KAIST AI.}  % leave blank if no need to mention equal contribution

\begin{abstract}
    Diffusion models have shown remarkable performance in generation problems over various domains including images, videos, text, and audio. A practical bottleneck of diffusion models is their sampling speed, due to the repeated evaluation of score estimation networks during the inference. In this work, we propose a novel framework capable of adaptively allocating compute required for the score estimation, thereby reducing the overall sampling time of diffusion models. We observe that the amount of computation required for the score estimation may vary along the time step for which the score is estimated. Based on this observation, we propose an early-exiting scheme, where we skip the subset of parameters in the score estimation network during the inference, based on a time-dependent exit schedule. Using the diffusion models for image synthesis, we show that our method could significantly improve the sampling throughput of the diffusion models without compromising image quality. Furthermore, we also demonstrate that our method seamlessly integrates with various types of solvers for faster sampling, capitalizing on their compatibility to enhance overall efficiency. The source code and our experiments are available at \url{https://github.com/taehong-moon/ee-diffusion}
\end{abstract}
\section{Introduction}
\label{main:sec:introduction}

Diffusion probabilistic models~\cite{sohl2015deep,ho2020denoising} have shown remarkable success in diverse domains including image synthesis~\citep{ho2020denoising,dhariwal2021diffusion,ho2022cascaded}, text-to-image generation~\citep{ramesh2022hierarchical,rombach2022high}, 3D point cloud generation~\citep{luo2021diffusion3d}, text-to-speech generation~\citep{jeong2021diff}, and video generation~\citep{ho2022video}. These models learn the reverse process of introducing noise into the data to data and denoise inputs progressively during inference using the learned reverse model.

One major drawback of diffusion models is their slow sampling speed, as they require multiple steps of forward passes through score estimation networks to generate a single sample, unlike the other methods such as GANs~\citep{goodfellow2014gan} that require only a single forward pass through a generator network. To address this issue, several approaches have been proposed to reduce the number of steps required for the sampling of diffusion models, for instance, by improving ODE/SDE solvers~\citep{kong2021fast,lu2022dpm,zhang2023fast} or distilling into models requiring less number of sampling steps~\citep{salimans2022progressive,song2023cm}.
Moreover, in accordance with the recent trend reflecting scaling laws of large models over various domains, 
diffusion models with a large number of parameters are quickly becoming mainstream as they are reported to produce high-quality samples~\citep{peebles2022scalable}. Running such large diffusion models for multiple sampling steps incurs significant computational overhead, necessitating further research to optimize calculations and efficiently allocate resources. 

On the other hand, recent reports have highlighted the effectiveness of early-exiting schemes in reducing computational costs for Large Language Models (LLMs)~\citep{schuster2022confident,hou2020dynabert,liu2021faster,schuster2021consistent}. The concept behind early-exiting is to bypass the computation of transformer blocks when dealing with relatively simple or confident words. Given that modern score-estimation networks employed in diffusion models share architectural similarities with LLMs, it is reasonable to introduce the early-exiting idea to diffusion models as well, with the aim of accelerating the sampling speed.

In this paper, we introduce Adaptive Score Estimation (ASE) for faster sampling from diffusion models, drawing inspiration from the early-exiting schemes utilized in LLMs. What sets diffusion models apart and distinguishes our proposal from a straightforward application of the early-exiting scheme is the time-dependent nature of the score estimation involved in the sampling process. We hypothesize that the difficulty of score estimation may vary at different time steps, and based on this insight, we adapt the computation of blocks differently for each time step. As a result, we gain the ability to dynamically control the computation time during the sampling procedure. To accomplish this, we present a time-varying block-dropping schedule and a straightforward algorithm for fine-tuning a given diffusion model to be optimized for this schedule. ASE successfully accelerates the sampling speed of diffusion models while maintaining high-quality samples. Furthermore, ASE is highly versatile, as it can be applied to score estimation networks with various backbone architectures and can be combined with different solvers to further enhance sampling speed. We demonstrate the effectiveness of our method through experiments on real-world image synthesis tasks.
\section{Related Work}
\paragraph{Fast Sampling of Diffusion Models.}
Diffusion probabilistic models~\citep{sohl2015deep,song2019generative,ho2020denoising,dhariwal2021diffusion} have shown their effectiveness in modeling data distributions and have achieved the state-of-the-art performance, especially in the field of image synthesis. 
These models employ a progressive denoising approach for noisy inputs which unfortunately lead to heavy computational costs.
To overcome this issue, multiple works have been proposed for fast sampling. DDIM~\citep{nichol2021improved} accelerates the sampling process by leveraging non-Markovian diffusion processes. FastDPM~\citep{kong2021fast} uses a bijective mapping between continuous diffusion steps and noises. DPM-Solver~\citep{lu2022dpm} analytically solves linear part exactly while approximating the non-linear part using high-order solvers. DEIS~\citep{zhang2023fast} utilizes exponential integrator and polynomial extrapolation to reduce discretization errors. In addition to utilizing a better solver, alternative approaches have been proposed, which involve training a student model using network distillation~\citep{salimans2022progressive}.
Recently, consistency model~\citep{song2023cm,song2023improved} proposed a distillation scheme to directly find the consistency function from the data point within the trajectory of the probability flow.
And \citet{kim2023consistency} refined the consistency model with input-output time parameterization within the score function and adversarial training.
While previous approaches focused on reducing the timestep of sampling, recent studies proposed an alternative way to accelerate sampling speed by reducing the processing time of diffusion model itself. In particular, Block Caching \citep{wimbauer2023cache} aim to re-use the intermediate feature which is already computed in previous timestep while Token Merging \citep{bolya2023token} target to reduce the number of tokens. Concurrent work \citep{tang2023deediff} suggests early exiting scheme on diffusion models. However, it requires additional module which is used to estimate an uncertainty of intermediate features.
Our work is orthogonal to these existing approaches, as we focus on reducing the number of processed blocks for each time step, rather than targeting a reduction in the number of sampling steps.

\paragraph{Early Exiting Scheme for Language Modeling.}
The recent adoption of Large Language Models (LLMs) has brought about significant computational costs, prompting interest in reducing unnecessary computations. Among the various strategies, an early-exiting scheme that dynamically selects computation layers based on inputs has emerged for Transformer-based LLMs. DynaBERT~\citep{hou2020dynabert} transfers knowledge from a teacher network to a student network, allowing for flexible adjustments to the width and depth. Yijin et al.~\citep{liu2021faster} employ mutual information and reconstruction loss to assess the difficulty of input words. CAT~\citep{schuster2021consistent} incorporates an additional classifier that predicts when to perform an early exit. CALM~\citep{schuster2022confident} constrains the per-token exit decisions to maintain the global sequence-level meaning by calibrating the early-exiting LLM using semantic-level similarity metrics. Motivated by the aforementioned works, we propose a distinct early-exiting scheme specifically designed for diffusion models.

\label{main:sec:related}
\section{Method}
This section describes our main contribution -  Adaptive Score Estimation (ASE) for diffusion models. The section is organized as follows. We first give a brief recap on how to train a diffusion model and provide our intuition on the time-varying complexity of score estimation. Drawing from such intuition, we empirically demonstrate that precise score estimation can be achieved with fewer parameters within a specific time interval. To this end, we present our early-exiting algorithm which boosts inference speed while preserving the generation quality.

\subsection{Time-Varying Complexity of Score Estimation}
\label{main:subsec:time_varying_complexity}

\paragraph{Training Diffusion Models.}
Let $x_0 \sim p_\text{data}(x) := q(x)$ be a sample from a target data distribution. In a diffusion model, we build a Markov chain that gradually injects Gaussian noises to $x_0$ to turn it into a sample from a noise distribution $p(x_T)$, usually chosen as standard Gaussian distribution. Specifically, given a noise schedule $(\beta_t)_{t=1}^T$, the forward process of a diffusion model is defined as

\[
q(x_t\given x_{t-1}) = \calN(x_t \given \sqrt{1-\beta_t} x_{t-1}, \beta_t I). 
\]

Then we define a backward diffusion process with a parameter $\theta$ as,
\[
 p_\theta(x_{1:T}) = p(x_T) \prod_{t=1}^T p_\theta(x_{t-1}\given x_t), \quad q(x_T\given x_0) \approx \calN(0, I).
\]

so that we can start from $x_T \sim \calN(0, I)$ and denoise it into a sample $x_0$. The parameter $\theta$ can be
optimized by minimizing the negative of the lower-bound on the log-evidence,
\[
\calL(\theta) &=
-\sum_{t=1}^T
\bbE_q
\left[
\KL[q(x_{t-1}\given x_t, x_0)\Vert p_\theta(x_{t-1}\given x_t)]
\right] \\ 
&\geq - \log p_\theta(x_0),
\]
where

\begin{equation}
    \begin{aligned}[b]
q(x_{t-1}|x_t,x_0) = \calN \left( x_{t-1} ; \tilde{\mu}_t(x_t,x_0), \tilde{\beta}_t I \right), \\
\tilde{\mu}_t(x_t,x_0)=\frac{1}{\sqrt{\alpha_t}}\left( x_t - \frac{\beta_t}{\sqrt{1-\bar{\alpha}_t}} \varepsilon_t \right).
    \end{aligned}
\end{equation}

The model distribution $p_\theta(x_{t-1}\given x_t)$ is chosen as a Gaussian,
\[
p_\theta(x_{t-1}\given x_t) = \calN(x_{t-1}\given \mu_\theta(x_t, t), \sigma^2_t I),\\
\mu_\theta(x_t, t) = \frac{1}{\sqrt{\alpha_t}}
\bigg( x_t - \frac{\beta_t}{\sqrt{1-\bar{\alpha}_t}}\varepsilon_\theta(x_t, t)\bigg),
\]
and the above loss function then simplifies to
\[
\calL(\theta) = \sum_{t=1}^T \bbE_{x_0, \varepsilon_t} \Big[ \lambda(t) \big\Vert \varepsilon_t - \varepsilon_\theta(\sqrt{\bar{\alpha}_t} x_0 + \sqrt{1-\bar\alpha_t}\varepsilon_t, t)\big\Vert^2\Big],
\]
where $\lambda(t) = \frac{\beta_t^2}{2\sigma_t^2\alpha_t(1-\bar\alpha_t)}$. The neural network $\varepsilon_\theta(x_t, t)$ takes a corrupted sample $x_t$ and estimates the noise that might have applied to a clean sample $x_0$.

Under a simple reparameterization, one can also see that,
\[
&\nabla_{x_t} \log q(x_t\given x_0) = -\frac{\varepsilon_t}{\sqrt{1-\bar\alpha_t}} \approx -\frac{\varepsilon_\theta(x_t,t)}{\sqrt{1-\bar\alpha_t}} := s_\theta(x_t, t),
\]
where $s_\theta(x_t, t)$ is the score estimation network. In this parameterization, the loss function can be written as,
\[
\calL(\theta) = \sum_{t=1}^T \bbE_{x_0, x_t}\Big[ \lambda'_t \Vert \nabla_{x_t}\log q(x_t\given x_0) - s_\theta(x_t, t)\Vert^2\Big],
\]
so learning a diffusion model amounts to regressing the score function of the distribution $q(x_t\given x_0)$. The optimal regressor of the score function $\nabla_{x_t} \log q(x_t)$ at time step $t$ is obtained by taking the expectation of the conditional score function over the noiseless distribution $\bbE_{x_0\given x_t}\left[\nabla_{x_t} \log q(x_t\given x_0)\right] = \nabla_{x_t} \log q(x_t)$.

Suppose we train our diffusion model using the standard parameterization (i.e., $\varepsilon$-parameterization), where the objective is to minimize the gap $\Vert \varepsilon_\theta - \varepsilon \Vert^2$. When $t$ is close to 1, this gap primarily represents noise, constituting only a small fraction of the entire $x_0$. Consequently, it indicates that learning does not effectively occur in the proximity to the noise. Given that a diffusion model is trained across all time steps with a single neural network, it is reasonable to anticipate that a significant portion of the parameters are allocated for the prediction of near data regime ($t$ close to 0). This intuition leads to our dropping schedule pruning more parameters when $t$ is close to 1.

\paragraph{Adaptive Computation for Score Estimation}
To get the samples from diffusion models, we can apply Langevin dynamics to get samples from the distribution given the score function $\nabla_{x}{\log p(x)}$. Depending on the number of iteration $N$ and step size $\beta$, we can iteratively update $x_{t}$ as follows:
\[
x_{t+1} = x_{t}+\beta\nabla_{x}\log p(x_{t})+\sqrt{2\beta}z_{t}, 
\]
where $z_{t} \sim \mathcal{N}(0, I)$.

Due to this iterative evaluation, the total sampling time can be roughly be computed as $T \times \tau$, where $T$ is the number of sampling steps and $\tau$ is the processing of diffusion model per time step. To enhance sampling efficiency, conventional approaches aim to reduce the number of time steps within the constrained value of $\tau$. Our experiments indicate that it's feasible to reduce $\tau$ by performing score estimation for specific time intervals using fewer parameters. While one could suggest employing differently sized models for estimating scores at various time intervals to reduce overall sampling time, our strategy introduces a simple early exiting framework within a single model, avoiding extra memory consumption.  Furthermore, our method focus on reducing the processing time $\tau$ while maintaining accurate predictions within a given time interval. To accomplish this, we introduce adaptive score estimation, wherein the diffusion model dynamically allocates parameters based on the time $t$. For challenging task such as time $t \rightarrow 0$, the full parameter is utilized, while it induces skipping the subset of parameters near prior distribution.

\subsection{Adaptive Layer Usage in Diffusion Process}
\label{main:subsec:early_exiting_diffusion}
We hereby introduce an early exiting framework to accelerate the sampling process of pre-trained diffusion models. Drawing upon the intuition presented in  \cref{main:subsec:time_varying_complexity}, we first explain how to decide the amount of parameters to be used for score estimation. After dropping the selected blocks, we design a fine-tuning algorithm to adjust the output of intermediate building blocks of diffusion models.

\begin{figure}[t]
\centering
\includegraphics[width=\linewidth]{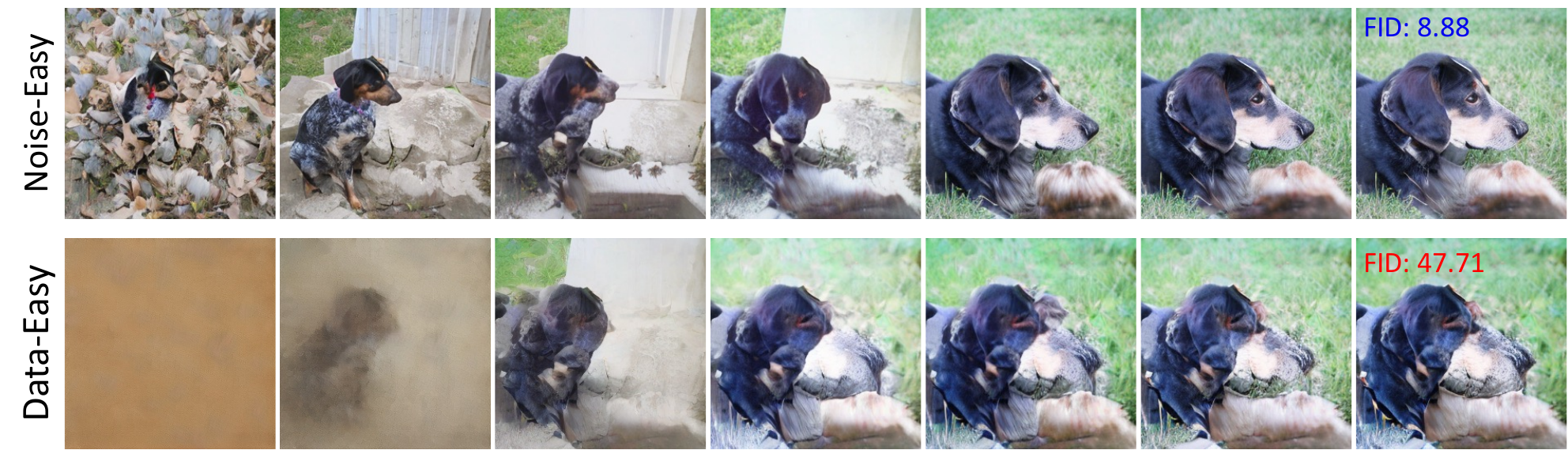}
\vspace*{-5mm}
\caption{Snapshot samples of Noise-Easy / Data-Easy schedules when fine-tuned DiT on ImageNet. 
While the data-easy schedule struggles to produce a discernible dog image, the noise-easy schedule successfully generates a clear dog image, achieving a converged FID score of 8.88.}
\label{fig:which_interval}
\end{figure}

\paragraph{Which time interval can be accurately estimated with fewer parameters?}
To validate our hypothesis in the context of training diffusion models, we conduct a toy experiment regarding the difficulty of score estimation for different time steps. We conduct tests under two scenarios: one assuming that estimation near the prior distribution requires fewer parameters (Noise-Easy schedule), and the other assuming that estimation near the data distribution demands fewer parameters (Data-Easy schedule). As shown in \cref{fig:which_interval}, one can easily find that the noise-easy schedule successfully generates a clear dog image where as the data-easy schedule struggles to produce a discernible dog image.

\begin{figure}[!t]
\centering
\includegraphics[width=\linewidth]{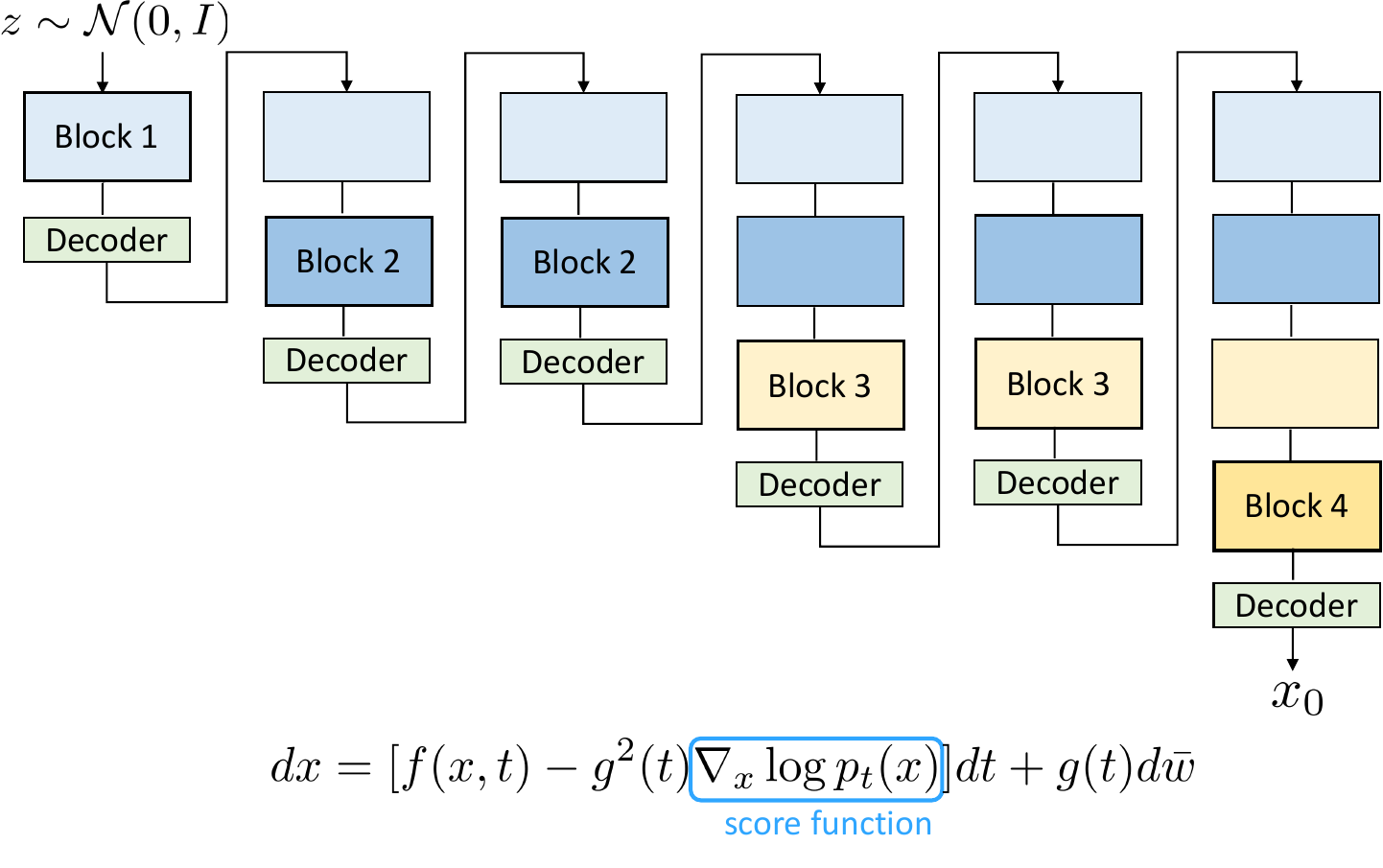}

\vspace*{-5mm}
\caption{Schematic for time-dependent exit schedule. Considering the varying difficulty of score estimation, we drop more building blocks of architecture near noise. While we skip the whole building blocks in DiT, we partially skip the blocks in U-ViT due to the long skip-connection.}
\label{fig:which_layer}
\end{figure}

\paragraph{Which layer can be skipped for score estimation?}
To accelerate inference in diffusion models, we implement a dropping schedule that takes into account the complexity of score estimation near $t\rightarrow 1$ compared to $t\rightarrow0$.
For the DiT model trained on ImageNet, which consists of 28 blocks, we design a dropping schedule that starts from the final block. Based on our intuition, we drop more DiT blocks as time approaches 1, as shown in \cref{fig:which_layer}. Conversely, for scores near the data, which represent more challenging tasks, we retain all DiT blocks to utilize the entire parameter set effectively.

In U-ViT, the dropping schedule has two main distinctions from DiT: the selection of candidate modules to drop and the subset of parameters to be skipped. Unlike DiT, we limit dropping to the decoder part in U-ViT. This decision is motivated by the presence of symmetric long skip connections between encoder and decoder, as dropping encoder modules induce the substantial information loss. Moreover, when dropping the parameters in U-ViT, we preserve the linear layer of a building block to retain feature information connected through skip connections, while skipping the remaining parameters.

\subsection{Fine-tuning Diffusion Models}
Following the removal of blocks based on a predetermined dropping schedule, we need to fine-tune the model. This is attributed to the early exit approach, where the intermediate outputs of each building block are directly connected to the decoder. Consequently, the decoder encounters input values that differ from the distribution it learned during its initial training, requiring adjustments.

To address this issue, we propose a novel fine-tuning algorithm that focuses on updating minimal information near time $t\rightarrow0$ while updating unseen information near time $t\rightarrow1$. To force the differential information update, we leverage two different techniques: (i) adapting Exponential Moving Average (EMA), and (ii) weighting the coefficients $\lambda(t)$.

The EMA technique is employed to limit the frequency of information updates, thereby preserving the previous knowledge acquired by the model during its initial training phase. A high EMA rate results in a more gradual modification of parameters. In our approach, we deliberately maintain a high EMA rate to enhance the stability of our training process. During the gradual parameter update, we aim to specifically encourage modifications in a subset of parameters that align the predicted scores more closely with the prior distribution. To prioritize the learning of this score distribution, we apply a higher coefficient to the $\lambda(t)$ term, which in turn multiplies on the expectation of the training loss. Once the model's performance appears to have plateaued, we adjust the $\lambda(t)$ value back to 1, aiming to facilitate comprehensive learning across the entire score distribution spectrum. We provide the pseudo-code for fine-tuning diffusion models in \cref{app:sec:experimental_details}.
\label{main:sec:methods}
\section{Experiments}
\label{main:sec:experiments}

\subsection{Experimental Setting}
\label{main:subsec:setting}

\paragraph{Experimental Details.} 
Throughout the experiments, we use DiT~\citep{peebles2022scalable} and U-ViT~\citep{bao2022all}, the two representative diffusion models. 
 
We employ three pre-trained models: (1) DiT XL/2 trained on ImageNet~\citep{krizhevsky2017imagenet} with the resolution of $256 \times 256$; (2) U-ViT-S/4 trained on CelebA~\citep{liu2015faceattributes} with the resolution of $64 \times 64$; (3) PixArt-$\alpha$-SAM-256 trained on SAM dataset~\citep{kirillov2023segment}. For the fine-tuning step in both DiT and U-ViT experiments, we employ a hybrid loss~\citep{nichol2021improved} with a re-weighted time coefficient and linear schedule for injecting noise. We use AdamW~\citep{loshchilov2017decoupled} optimizer with the learning rate of $2\cdot10^{-5}$. We use cosine annealing learning rate scheduling to ensure training stability for the U-ViT models. Batch size is set to 64, and 128 for fine-tuning DiT XL/2, U-ViT-S/4, respectively. We use $T=1000$ time steps for the forward diffusion process. In case of PixArt experiment, we fine-tune our model with 100K SAM data, the batch size of 200$\times$4, and 2200 iterations while the pre-trained model is trained with 10M data, the batch size of 176$\times$64 and 150K iterations. For further experimental details, we refer readers to \cref{app:sec:experimental_details}.

\paragraph{Evaluation Metrics.}
We employ Fréchet inception distance (FID)~\citep{heusel2017gans} for evaluating image generation quality of diffusion models. We compute the FID score between 5,000 generated samples from diffusion models and the full training dataset. In case of text-to-image experiment, we measure the FID score with MS-COCO valid dataset~\citep{lin2014microsoft}. To evaluate the sampling speed of diffusion models, we report the wall-clock time required to generate a single batch of images on a single NVIDIA A100 GPU.

\begin{figure*}[t]
    \centering
    \includegraphics[width=0.47\linewidth]{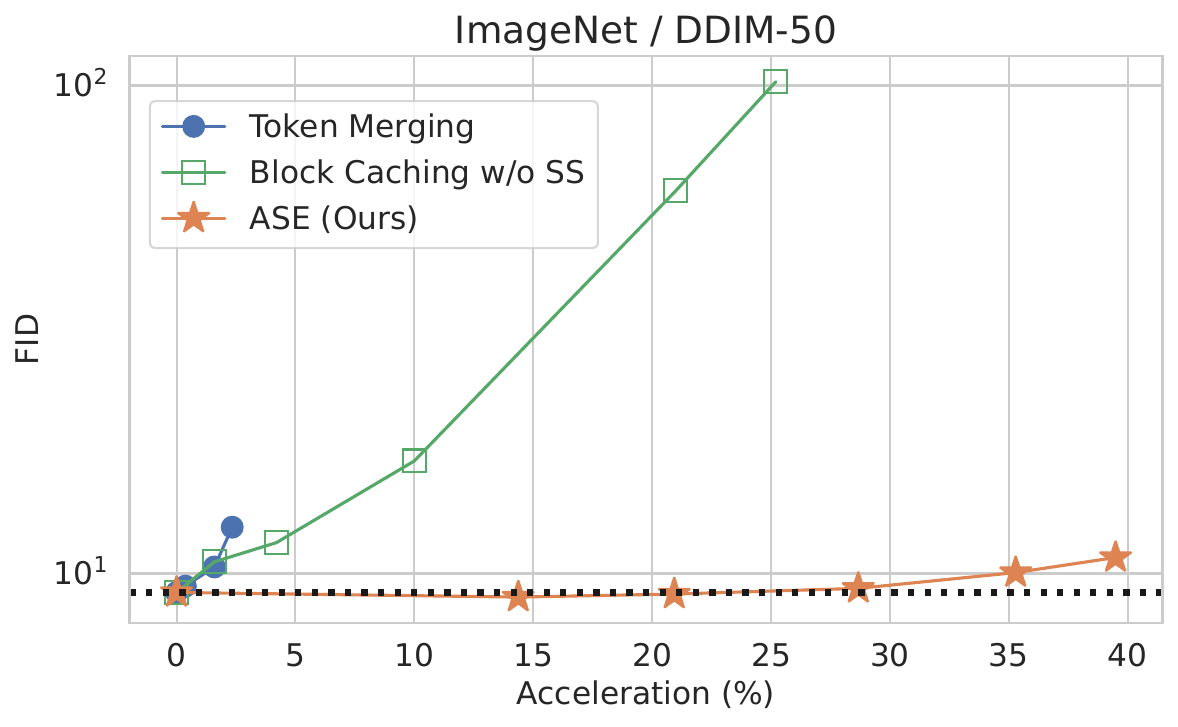}\quad
    \includegraphics[width=0.47\linewidth]{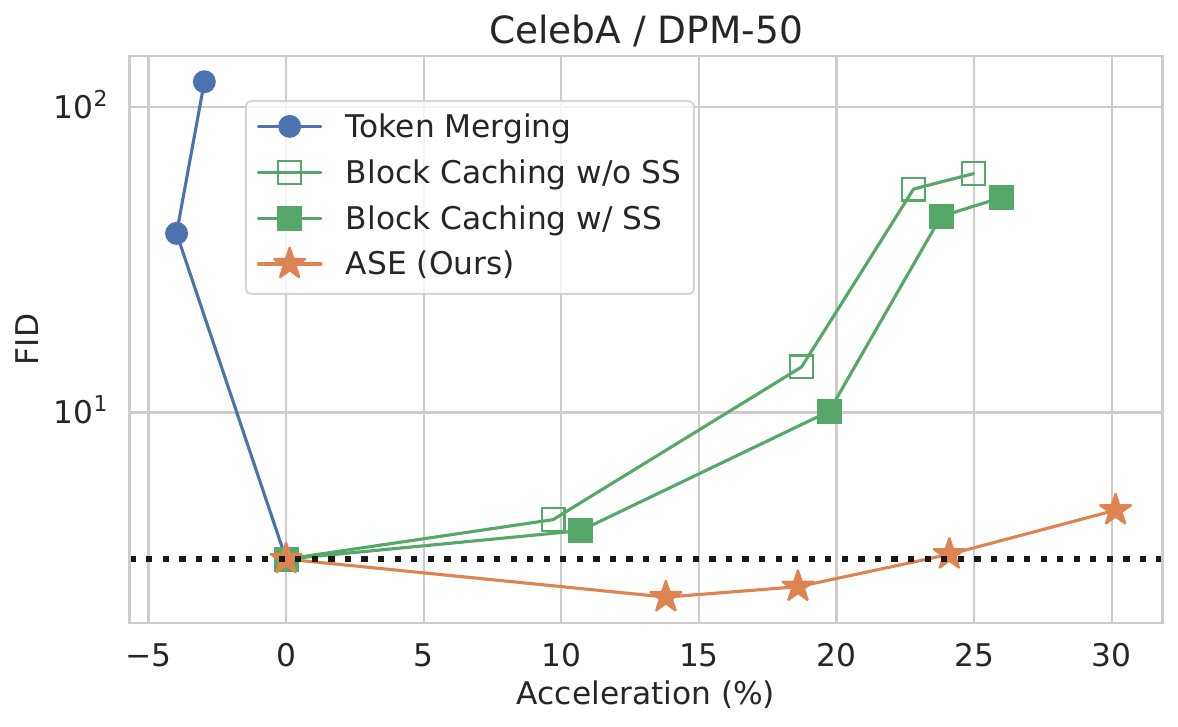}
    \vspace*{-2mm}
    \caption{Trade-off between image generation quality and sampling speed on ImageNet with DiT (left) and CelebA with U-ViT (right). We generate samples from DDIM and DPM sampler with $50$ steps for ImageNet and CelebA, respectively. ASE largely outperforms other techniques, preserving FID score while boosting sampling speed by approximately 25-30\%. Here, SS stands for scale-shift adjustment used together with block caching.
    }
    \label{fig:figure1}
\end{figure*}

\paragraph{Baselines.}
In this study, we benchmark our method against a range of recent techniques which aims reducing the processing time of diffusion models. This includes DeeDiff~\citep{tang2023deediff}, token merging~\citep[ToMe;][]{bolya2023token}, and block caching~\citep{wimbauer2023cache}. 
When extending ToMe to U-ViT architecture, we specifically apply the token merging technique to self-attention and MLP modules within each block of the U-ViT. Of note, U-ViT treats both time and condition as tokens in addition to image patches. To improve generative modeling, we exclude these additional tokens and focus solely on merging tokens associated with image patches, following the approach outlined by \citep{bolya2023token}. For block caching, we employ caching strategies within the attention layers. 
Naive caching may aggravate feature misalignment especially when caching is more aggressive in order to achieve faster sampling speed. To resolve such an issue, \citep{wimbauer2023cache} further propose shift-scale alignment mechanism. As we explore high-acceleration regime, we report results for both the original block caching technique and its variant with the shift-scale mechanism applied (termed SS in \cref{fig:figure1}). We only report the best performance attained among the diverse hyperparameter settings in the following sections. The remaining results will be deferred to \cref{app:sec:baselines} as well as experimental details for baseline strategies.

\begin{table*}[t]
    \renewcommand{\arraystretch}{1.2}
    \setlength{\tabcolsep}{1.5em}
    \centering
    
    \caption{
    Trade-off between image generation quality and sampling speed on ImageNet (DiT; DDPM sampler) and CelebA (U-ViT; EM sampler). ASE consistently maintains image generation quality while achieving a notable increase in sampling speed of approximately 30\%; ASE can be effectively used in conjunction with fast solvers. Refer to \cref{tab:detailed_schedule} in \cref{app:sec:experimental_details} for detailed description of our dropping schedules.
    }
    
    \vspace*{1mm}
    \parbox{0.35\linewidth}{%
        \resizebox{\linewidth}{!}{%
            \begin{tabular}{ccc}
                \toprule
                \multirow{2}{*}{\Centerstack{ImageNet\\(DiT)}}
                 & \multicolumn{2}{c}{DDPM-$250$}
                 \\
                \cmidrule(lr){2-3}
                 & \dmet{FID}     & \umet{Accel.} \\
                \midrule
                Baseline      &         9.078  & -       \\ 
                \midrule
                D2-DiT   &         8.662  & 23.43\% \\
                D3-DiT   & \textbf{8.647} & 30.46\% \\
                D4-DiT   &         9.087  & 34.56\% \\
                D7-DiT   &         9.398  & 38.92\% \\
                \bottomrule
            \end{tabular}
        }
    }
    \hspace{7mm}
    \parbox{0.35\linewidth}{%
        \resizebox{\linewidth}{!}{%
            \begin{tabular}{ccc}
                \toprule
                \multirow{2}{*}{\Centerstack{CelebA\\(U-ViT)}}
                 & \multicolumn{2}{c}{EM-$1000$}
                 \\
                \cmidrule(lr){2-3}
                 & \dmet{FID}     & \umet{Accel.} \\
                \midrule
                Baseline    &         2.944  &  -      \\ 
                \midrule
                D1-U-ViT & \textbf{2.250} & 21.3\% \\
                D2-U-ViT &         2.255  & 24.8\% \\
                D3-U-ViT &         3.217  & 29.7\% \\
                D6-U-ViT &         4.379  & 32.6\% \\
                \bottomrule
            \end{tabular}
        }
    }
    \label{tab:table1}
\end{table*}

\subsection{Inference Speed and Performance Trade-off}
Figure~\ref{fig:figure1} presents a trade-off analysis between generation quality and inference speed, comparing our approach to other baseline methods. We can readily find that ASE largely outperforms both ToMe and block caching strategies. ASE boosts sampling speed by approximately 25-30\% while preserving the FID score.

Techniques based on feature similarity, such as ToMe and block caching, are straightforward to implement yet fail to bring significant performance gain, or even in some cases, bring an increase in processing time. This can primarily be attributed to the additional computational overhead introduced by token partitioning and the complexity of bipartite soft matching calculations for token merging, which outweighs the advantages gained from reducing the number of tokens. This observation is particularly noteworthy, as even for the CelebA dataset, the number of tokens in U-ViT remains relatively small, and U-ViT does not decrease the token count through layers, as is the case with U-Net.

Regarding block caching, it yields only slight enhancements in inference speed while preserving the quality of generation. Although block caching can be straightforwardly applied to various diffusion models, it encounters a notable constraint: it relies significantly on scale-shift alignment, necessitating extra fine-tuning. Additionally, its effectiveness depends on the specifc architectural characteristics of the model being used. We postulate that this dependency may be related to the presence of residual paths within the architecture. It is crucial to highlight that our method effectively increases sampling speed without sacrificing the quality of the generated output.
 
 In \cref{tab:table2}, we further compare DeeDiff with our method using the performances reported in \citep[Table 1;][]{tang2023deediff}. ASE and DeeDiff share the same essence as both are grounded in the early-exiting framework. The distinction lies in the dynamic sampling process. To determine when to perform early-exiting for dynamic sampling, an additional module needs to be added to the model, whereas ASE does not require any additional memory. Furthermore, ASE exhibits faster acceleration while maintaining or improving FID, but for DeeDiff, there is a trade-off between the advantage in GFLOPs and the potential disadvantage in generation quality. In the case of ToMe and block caching, both methods fall significantly short of achieving the performance of ASE or DeeDiff. 

\subsection{Compatability with Diverse Sampling Solvers}
\begin{figure*}[t]
    \centering
    \includegraphics[width=0.85\linewidth]{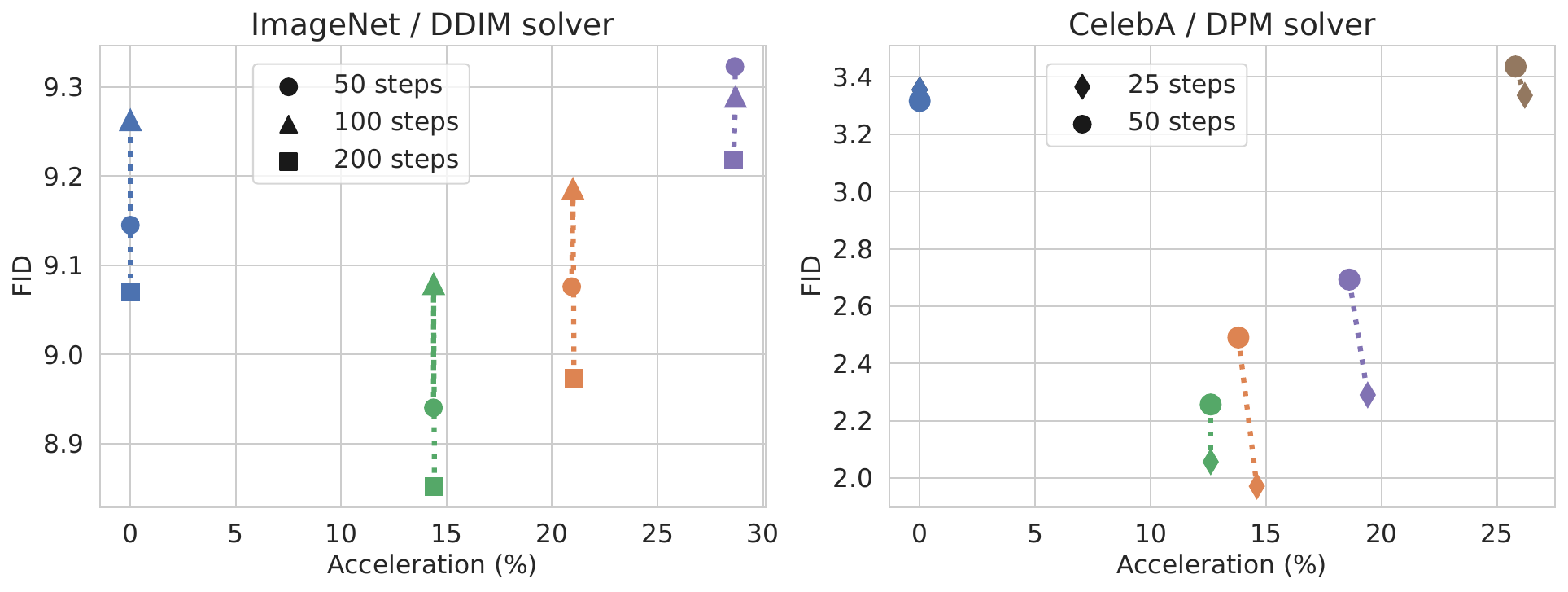}
    
    \vspace*{-3mm}
    \caption{
        Robustness of ASE across varying sampling timesteps: ImageNet with DDIM solver (left), and CelebA with DPM solver  (right). Both experiments employed U-ViT architecture. ASE displays robust performance throughout different timesteps in both different experimental settings. 
    }
    \label{fig:robustness}
\end{figure*}
We demonstrate the compatibility of the proposed method with diverse sampling methods. First of all, we verify that our method can be successfully applied to accelerate sampling speed without degrading generation qualtiy. In \cref{tab:table1}, we generated samples with DDPM \citep{ho2020denoising} in DiT architecture and get samples from Euler-Maruyama solver. Here, we present results of four varying dropping schedules in each experiments. In a nutshell, $n$ in D-$n$ schedule represents the acceleration scale. For instance, D3-DiT and D3-U-ViT schedules bring similar scales in terms of acceleration in sampling speed. We refer readers to \cref{tab:detailed_schedule} for detailed guide on ASE dropping schedules. 

Furthermore, we show that our method can be seamlessly incorporated with fast sampling solver, such as DDIM \citep{song2020denoising} solvers and DPM solver \citep{lu2022dpm}. From the DiT results presented in , we we observe that our approach effectively achieves faster inference while utilizing fewer parameters, yet maintains the same level of performance. In case of U-ViT, we show that our method notably achieves an over $30\%$ acceleration, while preserving similar quality in generation with the DPM solver. Notably in \cref{fig:robustness}, we highlight that our method is robust across various time steps within both DDIM and DPM solver. This indicates that our method effectively estimates scores across the entire time interval. The reasons for our method's robustness and efficiency in achieving faster inference will be further explained in \cref{main:sec:analysis}.

\subsection{Large Scale Text-to-Image Generation Task}
\begin{figure}[!ht]
    \centering
    \includegraphics[width=0.80\linewidth]{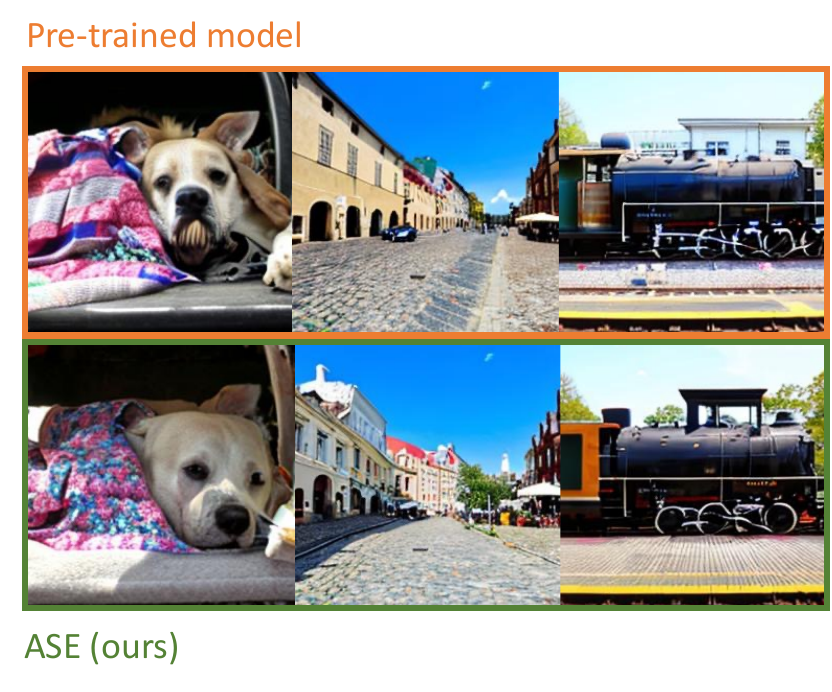}
    
    \vspace*{-4mm}
    \caption{
        Comparison between samples produced by pre-trained PixArt-$\alpha$ and ASE-enhanced PixArt-$\alpha$. Text prompts are randomly chosen.
    }
    \label{fig:pixart}
\end{figure}

\begin{table}[t]
    \caption{Trade-off between image generation quality and sampling speed on CelebA (U-ViT; DPM-50). Compared to the other baselines, ASE displays a remarkable sampling speed in terms of acceleration in GFLOPs.}
    
    \vspace*{1mm}
    \centering
    \resizebox{\columnwidth}{!}{
    \begin{tabular}{ccc}
        \toprule
        & \multicolumn{2}{c}{CelebA} \\
        \cmidrule(lr){2-3} 
        Methods  & \umet{Accel.} & \dmet{FID} \\
        \midrule
        U-ViT &    -     & 2.87       \\
        \midrule
        DeeDiff~\citep{tang2023deediff}   & 45.76\%  & 3.9   \\
        ToMe~\citep{bolya2023token}   & 3.05\%  & 4.963  \\
        Block Caching~\citep{wimbauer2023cache}   & 9.06\%  & 3.955   \\
        \midrule
        ASE (Ours) & 23.39\%  & 1.92  \\
        \bottomrule
    \end{tabular}
    }
    \label{tab:table2}
\end{table}
To demonstrate that our method can be extended to large-scale datasets, we apply it to the pre-trained PixArt-$\alpha$ model. While there may be concerns that fine-tuning with a large-scale dataset could potentially slow down the fine-tuning process, we find that using only 1$\%$ of the original data is sufficient for our method to achieve the desired performance. To evaluate our method, we employ a DPM solver with 20 steps and classifier-free guidance~\citep{ho2022classifier}. Although the original model achieves an FID score of 12.483, the ASE-enhanced model attains an FID score of 12.682, with a 14$\%$ acceleration in terms of wall-clock time. An example of an image generated from a given prompt is shown in \cref{fig:pixart}.
\section{Further Analysis}
\label{main:sec:analysis}
\begin{figure}[!t]
    \centering
    \includegraphics[width=0.8\linewidth]{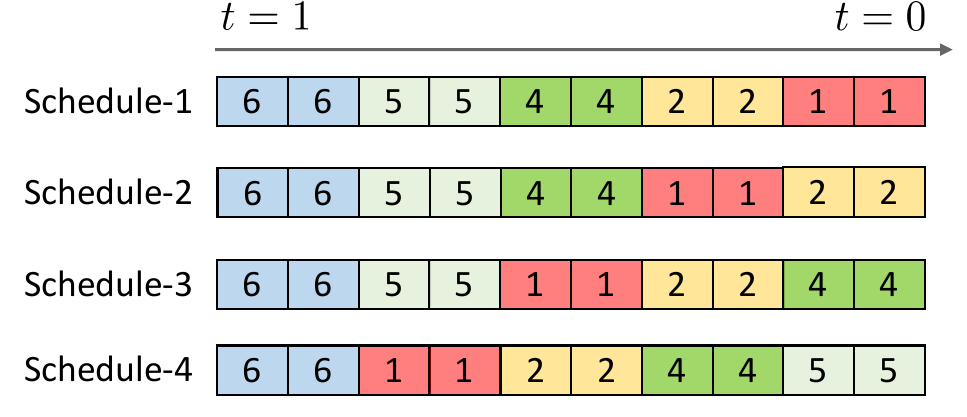}
    \vspace*{-3mm}
    \caption{Dropping schedules designed for the ablation study. We divide the sampling time into ten uniform intervals, and drop a specific amount of blocks. The number indicates the amount of blocks left after dropping the rest.}
    \label{fig:figure6}
\end{figure}

\paragraph{Ablation Study on Dropping Schedules.} 
Although it is empirically understood that we can eliminate more parameters near the prior distribution, it remains to be determined which time-dependent schedules yield optimal performance in generation tasks. To design an effective dropping schedule, we conduct an ablation study as follows: we create four distinct schedules that maintained the same total amount of parameter dropping across all time intervals, but vary the amount of dropping for each specific interval. These schedules are tested on a U-ViT backbone trained on the CelebA dataset. Specifically, the decoder part of this architecture consists of six blocks, and \cref{fig:figure6} illustrates how many blocks are utilized at each time $t$. By fine-tuning in this manner, we evaluate the generation quality of the models, as shown in \cref{tab:table3}. As the results indicate, Schedule 1 outperforms the others, demonstrating the most superior and stable performance across varying time steps.

\paragraph{Viewpoint of Multi-task Learning.}
Diffusion models can be seen as a form of multi-task learning, as they use a single neural network to estimate the scores at every time \( t \). In the context of multi-task learning, negative transfer phenomenon can occur, leading to a decrease in the generation quality of diffusion models. Recent work, such as DTR~\citep{park2023dtr}, improve generation quality by jointly training a mask with the diffusion model. This approach minimizes negative transfer by reducing interference between tasks. Similarly, our method, despite using fewer parameters, is designed to achieve a comparable effect. By explicitly distinguishing the parameters used for predicting specific intervals through early-exiting, our approach can mitigate the issues associated with negative transfer. 

\begin{table}[!t]
    \caption{FID score on CelebA dataset with U-ViT backbone across ablated dropping schedules. In both DPM-25 and DPM-50, schedule-1 exhibits the best performance.}
    
    \vspace*{1mm}
    \label{tab:table3}
    \centering
    \resizebox{0.55\linewidth}{!}{%
         \begin{tabular}{cccc}
         \toprule
            Methods  & DPM-25  & DPM-50 \\
            \midrule
            Schedule-1  & {\bf 2.116} & {\bf 2.144} \\
            Schedule-2  & 2.456 & 2.28 \\
            Schedule-3  & 2.173 & 3.128 \\
            Schedule-4  & 2.966 & 3.253 \\
            \bottomrule
        \end{tabular}
    }
\end{table}
\begin{figure}[t]
    \centering
    \includegraphics[width=0.9\linewidth]{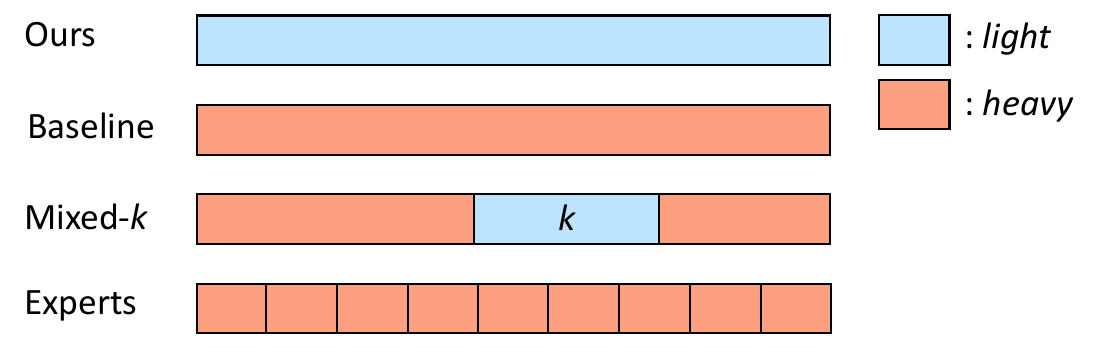}
    
    \vspace*{-3mm}
    \caption{Schematic for different types of dropping schedules designed to validate negative transfer phenomenon. Mixed-k replaces the original heavy model with light ASE model only on $k$th time interval. Experts employ individually fine-tuned heavy models at each time interval.}
    \label{fig:figure7}
\end{figure}

\begin{table}[t]
    \caption{FID score on CelebA dataset with U-ViT backbone across NTR-inspired dropping schedules. Experts outperform both baseline and further fine-tuned model thereby indicating that negative transfer does exist. Moreover, all the mixed-k schedules, despite only replacing a single time interval, demonstrate improved performance compared to the original baseline model.}
    \vspace*{1mm}
    \label{tab:table4}
    \centering
    \resizebox{0.6\linewidth}{!}{
         \begin{tabular}{ccc}
         \toprule
            Methods  & DPM-25  & DPM-50 \\
            \midrule
            Baseline   & 3.355 & 3.316 \\
            Further-trained & 4.262 & 4.028 \\
            Multi-Experts   & {\bf 2.987} & {\bf 2.942} \\
            \midrule
            Mixed-1  & 2.938 & 3.054 \\
            Mixed-3  & 2.654 & 3.232 \\
            Mixed-5  & 3.287 & 3.187 \\
            Mixed-7  & 2.292 & 2.969 \\
            Mixed-9  & 2.933 & 3.027 \\
            \bottomrule
        \end{tabular}
    }
\end{table}
To illustrate the efficacy of our method in mitigating negative transfer, we hereby conduct a toy experiment. Consider score estimation over a specific time interval $t \in [s, l]$ as a single task. In the experiment, we equally divide the whole sampling time into ten intervals, thereby defining a total of ten tasks. To verify the presence of negative transfer in the diffusion model, we create both a baseline model and expert models trained specifically for each interval. In order to check whether the pre-trained model is sufficiently trained, we further train the baseline model, and \cref{tab:table4} shows that further-training degrades the performance. Also, the multi-experts model outperforms the baseline model, indicating successful reduction of task interference. Furthermore, replacing the pre-trained model with the ASE module (\textit{Mixed-\(k \)} models) in a single time interval leads to performance gains. In \cref{tab:table4}, we can readily observe that the mixed schedules outperform the baseline model across all intervals in terms of image generation quality. This finding suggests that our training approach can not only effectively boost sampling speed but also preserves model performance via mitigating negative transfer effect.
\section{Conclusion and Limitations}
\label{main:sec:conclusion}
In this paper, we present a novel method that effectively reduces the overall computational workload by using an early-exiting scheme in diffusion models. Specifically, our method adaptively selects the blocks involved in denoising the inputs at each time step, taking into account the assumption that fewer parameters are required for early denoising steps. 

Surprisingly, we demonstrate that our method maintains performance in terms of FID scores even when reducing calculation costs by 30\%. Our approach is not limited to specific architectures, as we validate its effectiveness on both U-ViT and DiTs models. A limitation of our proposed method is that we manually design the schedule for the early-exiting scheme. As future work, we acknowledge the need to explore automated methods for finding an optimal schedule.

\section*{Impact Statement} % Societal Impact
Our work is improving diffusion models which can be misused for generating fake images or videos, contributing to the spread of deepfake content or the creation of misleading information. Also, given that these models are trained on data collected from the internet, there is a risk of harmful biases being embedded in the generated samples such as emphasizing stereotypes.

\section*{Acknowledgement}
% [Version 1]
The authors would like to express their sincere gratitude to Jaehyeon Kim and Byeong-Uk Lee for their insightful and constructive discussions. This work was partly supported by Institute for Information \&
communications Technology Promotion(IITP) grant funded by the Korea government(MSIT)
(No.RS-2019-II190075 Artificial Intelligence Graduate School Program(KAIST), KAIST-NAVER Hypercreative AI Center, Korea Foundation for Advanced Studies (KFAS), No.2022-0-00713, Meta-learning Applicable to Real-world Problems), and National Research Foundation of Korea (NRF) funded by the Ministry of Education (NRF2021M3E5D9025030).

\bibliography{references}
\bibliographystyle{icml2024}

% \clearpage
% \newpage
\appendix
\onecolumn
\section{Experimental Details}
\label{app:sec:experimental_details}

\subsection{How to design dropping schedules?}
\paragraph{Diverse Time-dependent Dropping Schedules}
In \cref{tab:table1}, we briefly introduce the difference between the diverse schedules, D1 to D6. We hereby provide the formal definition of D-$n$ schedules. We refer the reader to \cref{tab:detailed_schedule}. First, the sampling time $[0,1]$ is divided into ten intervals with equal length.

For the DiT architecture, we designated the blocks to be dropped among the total of 28 blocks. In the case of D1-DiT, we utilized all 28 blocks near the data. As we moved towards the noise side, we gradually discarded some blocks per interval, resulting in a final configuration of using the smallest number of blocks near the noise. The higher the number following 'D', the greater the amount of discarded blocks, thereby reducing the processing time of the diffusion model. For the most accelerated configuration, D7-DiT, we designed a schedule where only 8 blocks pass near the noise.

\begin{table*}[!hb]
\renewcommand{\arraystretch}{1.1}
\caption{Number of blocks used for varying dropping schedules. All schedules use the same number of blocks within a fixed time interval. Of note, $n$ in D-$n$ schedule represents the acceleration scale. For instance, D3-DiT and D3-U-ViT schedules bring similar scales in terms of acceleration in sampling speed. Reported acceleration performance is measured with DDPM and EM solver applied to DiT and U-ViT, respectively.}
\label{tab:detailed_schedule}
\centering
\resizebox{0.95\linewidth}{!}{%
\begin{tabular}{cccccccccccc}
    \toprule
    \multirow{2}{*}{Schedule} & \multirow{2}{*}{Acceleration} 
 & \multicolumn{5}{c}{Sampling timestep $t$} \\
    \cmidrule(lr){3-12} 
     & & $[0,0.1]$ & $[0.1, 0.2]$ & $[0.2, 0.3]$ & $[0.3, 0.4]$ & $[0.4,0.5]$ & $[0.5,0.6]$ & $[0.6,0.7]$ & $[0.7,0.8]$ & $[0.8,0.9]$ & $[0.9,1.0]$ \\
    \midrule
    D2-DiT & 23.43\% & 28 & 28 & 25 & 25 & 22 & 22 & 19 & 19 & 16 & 16 \\
    D3-DiT & 30.46\% & 28 & 28 & 24 & 24 & 20 & 20 & 16 & 16 & 12 & 12 \\
    D4-DiT & 34.56\% & 28 &	28 & 26 & 24 & 20 & 18 & 12 & 10 & 8 & 8 \\
    D7-DiT & 38.92\% & 28 &	28 & 24	& 21 & 18 &	15 & 10	& 10 & 8 & 8 \\
    \midrule
    D1-U-ViT & 21.3\% & 6 &	6 &	4 &	4 &	2 &	2 &	2 &	2 & 1 & 1 \\
    D2-U-ViT & 24.8\% & 5 &	5 &	4 &	4 &	2 &	2 &	1 &	1 & 1 &	1 \\
    D3-U-ViT & 29.7\% & 3 &	3 &	2 &	2 &	2 &	2 &	1 &	1 & 1 &	1 \\
    D6-U-ViT & 32.6\% & 2 &	2 &	2 &	2 &	1 &	1 &	1 & 1 & 1 &	1 \\
    \bottomrule
\end{tabular}}
\end{table*}

\begin{figure*}[!hb]
\centering
\includegraphics[width=.7\linewidth]{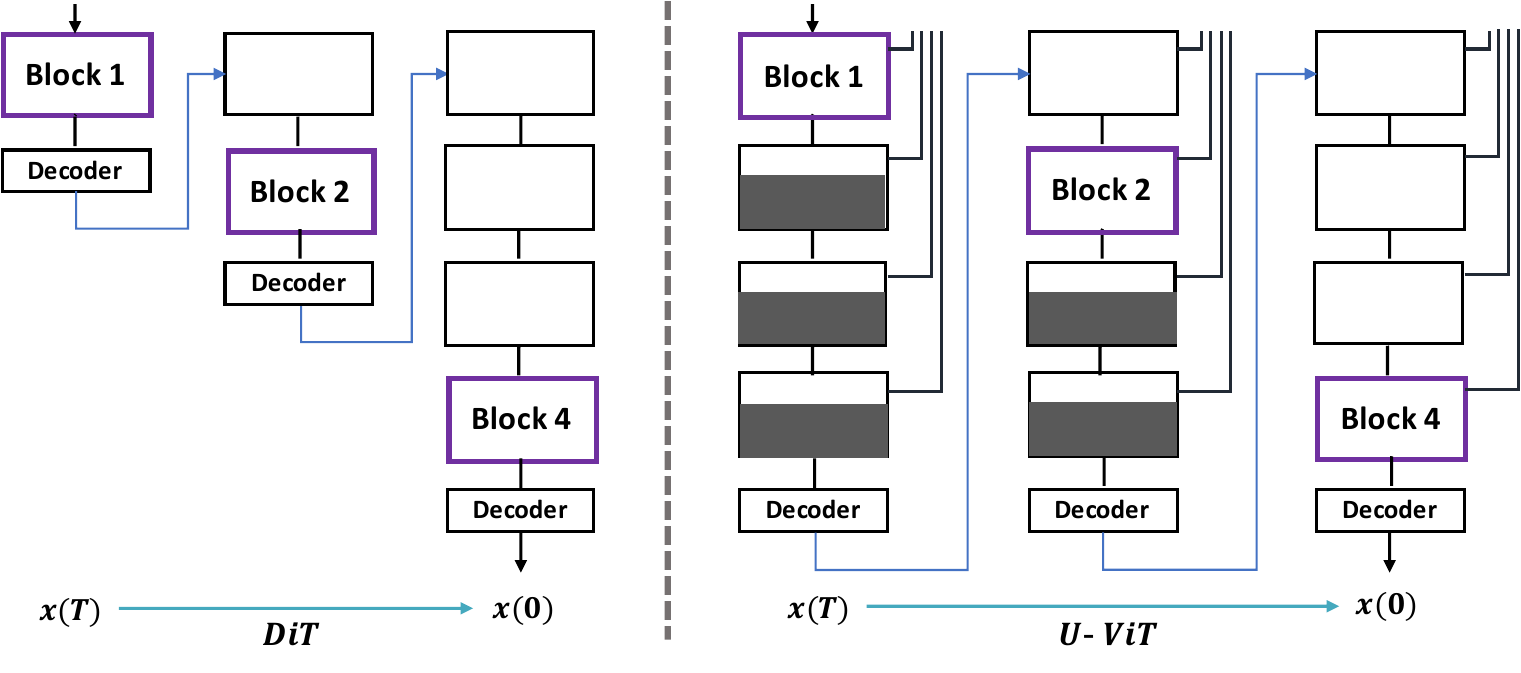}
\caption{Schematic for the dropping schedules of DiT (left) and U-ViT (right). Due to the existence of residual connections in U-ViT, dropping encoder or decoder blocks in a straightforward manner cause severe performance degradation. In the case of U-ViT, the decoder blocks, except for the linear layer connected to encoder residual connections, are dropped.}
\label{fig:schematic_u_vit}
\end{figure*}

For the U-ViT architecture as we depicted in \cref{fig:schematic_u_vit}, we aimed to preserve the residual connections by discarding sub-blocks other than nn.Linear, rather than skipping the entire building block. Additionally, the target of dropping was limited to the decoder part, distinguishing it from DiT. Similarly, for D1-U-ViT, we allowed the entire decoder consisting of 6 blocks to pass near the data, and as we moved towards the noise side, we gradually discarded a single block per interval, resulting in only 1 blocks passing near the noise, while the remaining blocks only passed through nn.Linear.

\subsection{Pseudo-code for fine-tuning diffusion models}
\begin{algorithm}[!ht]
\caption{Adjusting the output of intermediate building block of diffusion models}
\label{alg:adjusting}
\begin{algorithmic}
    \REQUIRE Training dataset $\calD$, Teacher parameter $\theta_{T} = [\theta_T^1, \dots, \theta_T^N]$, Student parameter $\theta_{S} = [\theta_S^1, \dots, \theta_S^N]$, EMA rate $\alpha$, Pre-defined Exit Schedule $S(t)$, Time-dependent coefficient $\lambda(t)$, Re-weighting cycle $C$, Learning rate $\eta$.
    
    \vspace{2mm}
    \STATE $\theta_{T} \gets \theta_{S}, t \sim [0, 1]$ 
    \WHILE{not converged}
    \STATE Sample a mini-batch $\calB \sim \calD$.
    \FOR{$i=1,\dots,|\calB|$}
        \STATE Take the input $\bsx_i$ from $\calB$.
        \FOR{$l=1,\dots,N$}
            \IF{$l \leq S(t)$}
                \STATE $\tilde{\bsx_i} \gets \text{perturb}(\bsx_i, t)$
                \STATE $\ell_i \gets \lambda(t) \cdot \text{loss}(\tilde{\bsx_i}, t)$ 
            \ELSE
                \STATE Break for loop
            \ENDIF
        \ENDFOR
    \ENDFOR
    \STATE $\theta_{S} \gets \theta_{S} - \eta \nabla_{\theta_{S}} \frac{1}{|\calB|}\sum_i\ell_i$.
    
    \vspace{2mm}
    \STATE Update $\theta_{T} \gets \alpha \theta_{T} + (1-\alpha) \theta_{S}$ 
    \ENDWHILE
\end{algorithmic}
\end{algorithm}

\subsection{Computational Efficiency of ASE}
\paragraph{Additional Fine-tuning cost of ASE}
Compared with ToMe~\cite{bolya2023token} and Block Caching~\citep{wimbauer2023cache}, our method requires fine-tuning. Nonetheless, we demonstrate its negligible fine-tuning cost and high efficiency by reporting the computational costs for fine-tuning in \cref{tab:table6}.
\begin{table*}[!hb]
    \renewcommand{\arraystretch}{1.2}
    \setlength{\tabcolsep}{1.5em}
    \centering
    \caption{
    Fine-tuning costs when we apply ASE into pre-trained DiT on ImageNet and U-ViT on CelebA. These tables show the number of iterations and batch sizes used during the fine-tuning process.
    }
    \parbox{0.30\linewidth}{%
        \resizebox{\linewidth}{!}{%
            \begin{tabular}{cc}
                \toprule
                \Centerstack{ImageNet\\(DiT)} & iteration * batch size \\
                \midrule
                Baseline    & 400K * 256 \\ 
                \midrule
                D2-DiT & 400K * 32 (12.50$\%$) \\
                D3-DiT & 450K * 32 (14.06$\%$) \\
                D4-DiT & 500K * 32 (15.63$\%$) \\
                \bottomrule
            \end{tabular}
        }
    }
    \hspace{7mm}
    \parbox{0.30\linewidth}{%
        \resizebox{\linewidth}{!}{%
           \begin{tabular}{c c}
                \toprule
                \Centerstack{CelebA\\(U-ViT)} & iteration * batch size \\
                \midrule
                Baseline    & 500K * 128 \\ 
                \midrule
                D1-U-ViT & 40K * 128 (8$\%$) \\
                D2-U-ViT & 50K * 128 (10$\%$) \\
                D3-U-ViT & 150K * 64 (15$\%$) \\
                D6-U-ViT & 200K * 64 (20$\%$) \\
                \bottomrule
            \end{tabular}
        }
    }
    \label{tab:table6}
\end{table*}

\paragraph{Results on actual inference time of ASE}
In \cref{tab:table7}, we provide additional results on wall-clock time. We note that the acceleration rate in the original paper is also measured in terms of wall-clock time.
\begin{table*}[!hb]
    \renewcommand{\arraystretch}{1.2}
    \setlength{\tabcolsep}{1.5em}
    \centering
    \caption{
   Wall-clock time of generating samples with ASE-enhanced models. Left table is the result of DiT model fine-tuned on ImageNet and right table is the result of U-ViT model fine-tuned on CelebA.
    }
    \parbox{0.35\linewidth}{%
        \resizebox{\linewidth}{!}{%
            \begin{tabular}{ccc}
                \toprule
                \multirow{2}{*}{\Centerstack{ImageNet\\(DiT)}}
                 & \multicolumn{2}{c}{DDPM-$250$}
                 \\
                \cmidrule(lr){2-3}
                 & \dmet{FID}     & \dmet{Wall-clock time (s)} \\
                \midrule
                Baseline      &         9.078  & 59.60       \\ 
                \midrule
                D2-DiT   &         8.662  & 45.63 \\
                D3-DiT   & \textbf{8.647} & 41.44 \\
                D4-DiT   &         9.087  & 39.00 \\
                D7-DiT   &         9.398  & 36.40 \\
                \bottomrule
            \end{tabular}
        }
    }
    \hspace{7mm}
    \parbox{0.35\linewidth}{%
        \resizebox{\linewidth}{!}{%
            \begin{tabular}{ccc}
                \toprule
                \multirow{2}{*}{\Centerstack{CelebA\\(U-ViT)}}
                 & \multicolumn{2}{c}{EM-$1000$}
                 \\
                \cmidrule(lr){2-3}
                 & \dmet{FID}     & \dmet{Wall-clock time (s)} \\
                \midrule
                Baseline    &         2.944  &  216.70      \\ 
                \midrule
                D1-U-ViT & \textbf{2.250} & 170.54 \\
                D2-U-ViT &         2.255  & 162.95 \\
                D3-U-ViT &         3.217  & 152.34 \\
                D6-U-ViT &         4.379  & 146.05 \\
                \bottomrule
            \end{tabular}
        }
    }
    \label{tab:table7}
\end{table*}

\section{Related Work}
\paragraph{Transformers in Diffusion Models.}
The pioneering diffusion models~\citep{ho2020denoising,song2019generative,dhariwal2021diffusion}, especially in the field of image synthesis, have adopted a U-Net~\citep{ronneberger2015u} backbone architecture with additional modifications including the incorporation of cross- and self-attention layers. Motivated by the recent success of transformer~\citep{vaswani2017attention} networks in diverse domains~\citep{brown2020language,devlin2019bert,xie2021segformer,strudel2021segmenter,liu2022video}, several studies have attempted to leverage the Vision Transformer (ViT)~\citep{dosovitskiy2021an} architecture for diffusion models.  Gen-ViT~\citep{yang2022your} is a pioneering work that shows that standard ViT can be used for diffusion backbone.  U-ViT~\citep{bao2022all} enhances ViT's performance by adding long skip connections and additional convolutional operation. Diffusion Transformers (DiTs)~\citep{peebles2022scalable} investigate the scalability of transformers for diffusion models and demonstrate that larger models consistently exhibit improved performance, albeit at the cost of higher GFLOPs. Our approach focuses on enhancing the efficiency of the transformer through adaptive block selection during calculations, and can be applied to existing transformer-based approaches, such as DiTs, to further optimize their performance.

\section{Further Analysis on Baselines}
\label{app:sec:baselines}

\paragraph{Analysis on ToMe}
In this section, we conducted experiments on three different cases for applying ToMe to the building block of a given architecture. The `F' schedule denotes applying ToMe starting from the front-most block, the `R' schedule denotes starting from the back-most block, and the `B' schedule represents symmetric application from both ends. In the \cref{fig:figure1}, we report the experiment results that showed the most competitive outcomes. Furthermore, we present the remaining experiments conducted using various merging schedules, as illustrated in \cref{tab:tome_dit}, \cref{tab:add_tome_u_vit_dpm_50}. In summary, for the DiT architecture, the `B' schedule performed well, while the `R' schedule demonstrated satisfactory performance for the U-ViT architecture.

\begin{table}[!hb]
\renewcommand{\arraystretch}{1.2}
\caption{Diverse merging schedule experiments on DiT with DDIM sampler.}
\centering
\resizebox{1.\linewidth}{!}{
\begin{tabular}{ccccccccccccc}
    \toprule
    \multirow{2}{*}{DDIM-50} & \multicolumn{2}{c}{B2} &
    \multicolumn{2}{c}{B4} & \multicolumn{2}{c}{B6} &
    \multicolumn{2}{c}{B8} & \multicolumn{2}{c}{All} \\
    \cmidrule(lr){2-3} \cmidrule(lr){4-5} \cmidrule(lr){6-7} \cmidrule(lr){8-9} \cmidrule(lr){10-11} \cmidrule(lr){12-13} 
     & \dmet{FID} & \umet{Accel.} & \dmet{FID} & \umet{Accel.} & \dmet{FID} & \umet{Accel.} & \dmet{FID} & \umet{Accel.} & \dmet{FID} & \umet{Accel.} \\
    \midrule
    attn-ratio-2-down-1  & 9.172 &  0.29\% & 9.421 & 0.37\% & 10.43 & 0.60\% & 13.926 & 0.69\% & 117.194 & 1.92\% \\ 
    attn-ratio-3-down-1  & 9.313 &  0.49\% & 9.745 & 0.82\% & 12.918 & 1.03\% & 22.495 & 1.45\% &  170.170 & 6.08\% \\ 
    attn-ratio-4-down-1  & 9.409 &  0.85\% & 10.314 & 1.59\% & 17.567 & 2.27\% & 37.763 & 2.97\% & 214.759 & 10.34\% \\ 
    attn-ratio-5-down-1  & 9.741 &  0.91\% & 11.284 & 2.26\% & 25.675 & 2.63\% & 58.550 & 4.07\% & 247.608 & 16.66\% \\ 
    attn-ratio-6-down-1  & 10.014 & 0.99\% & 12.441 & 2.34\% & 38.124 & 3.72\% & 81.987 & 5.07\% & 274.591 & 21.55\% \\ 
    \bottomrule
\end{tabular}
\label{tab:tome_dit}
}
\end{table}

\begin{table*}[!hb]
\renewcommand{\arraystretch}{1.2}
\caption{Diverse merging schedule experiments on U-ViT with DPM sampler.}
\centering
\resizebox{0.8\linewidth}{!}{%
\begin{tabular}{ccccccccccccc}
    \toprule
    \multirow{2}{*}{DPM-50} & \multicolumn{2}{c}{R2} &
    \multicolumn{2}{c}{R3} & \multicolumn{2}{c}{R4} &
    \multicolumn{2}{c}{R5} \\
    \cmidrule(lr){2-3} \cmidrule(lr){4-5} \cmidrule(lr){6-7} \cmidrule(lr){8-9} \cmidrule(lr){10-11}
     & \dmet{FID} & \umet{Accel.} & \dmet{FID} & \umet{Accel.} & \dmet{FID} & \umet{Accel.} & \dmet{FID} & \umet{Accel.} \\
    \midrule
    attn-ratio-2-down-1  & 38.505 &  -3.98\% & 45.544 & -5.89\% & 65.755 & -7.51\% & 79.086 & -9.15\%  \\ 
    attn-ratio-3-down-1  & 120.596 &  -2.97\% & 141.073 & -4.53\% & 200.132 & -5.85\% & 232.040 & -7.07\%  \\ 
    attn-ratio-4-down-1  & 264.153 &  -2.13\% & 279.270 & -2.76\% & 311.823 & -3.69\% & 319.599 & -4.57\%  \\ 
    attn-ratio-5-down-1  & 308.350 &  -1.13\% & 315.334 & -1.53\% & 332.565 & -1.90\% & 343.486 & -2.02\% \\ 
    attn-ratio-6-down-1  & 330.501 & 0.05\% & 344.353 & 0.41\% & 362.002 & 0.69\% & 372.612 & 1.10\% \\ 
    \bottomrule
\end{tabular}
\label{tab:add_tome_u_vit_dpm_50}
}
\end{table*}

\paragraph{Analysis on Block Caching}
To ensure fair comparison between baseline methods, we faithfully implement block caching algorithm on both DiT and U-ViT architecture. In this experiment, we applied it to the attention part of the U-ViT blocks, and \cref{tab:add_block_caching_u_vit_dpm_50} shows the trade-off between generation quality and inference speed depending on the presence or absence of the scale-shift mechanism.

\begin{table*}[!hb]
\renewcommand{\arraystretch}{1.2}
\caption{Additional block caching experiments on U-ViT with DPM sampler.}
\centering
\resizebox{0.4\linewidth}{!}{%
\begin{tabular}{cccccc}
    \toprule
    \multirow{2}{*}{DPM-50} & \multicolumn{2}{c}{Attn(wo SS)} &
    \multicolumn{2}{c}{Attn(w SS)}  \\
    \cmidrule(lr){2-3} \cmidrule(lr){4-5}
     & \dmet{FID} & \umet{Accel.} & \dmet{FID} & \umet{Accel.}  \\
    \midrule
    attn-ths-0.1  & 4.462 &  9.70\% & 3.955 & 9.06\%  \\ 
    attn-ths-0.2  & 14.083 &  18.73\% & 9.707 & 18.11\% \\ 
    attn-ths-0.3  & 53.770 &  22.80\% & 32.518 & 22.35\% \\ 
    attn-ths-0.4  & 60.390 &  24.98\% & 45.523 & 24.26\% \\ 
    \bottomrule
\end{tabular}
\label{tab:add_block_caching_u_vit_dpm_50}
}
\end{table*}
\section{Qualitative Comparison}
\label{app:sec:additional_experiments}
\begin{figure*}[!hb]
\centering
\includegraphics[width=0.45\linewidth]{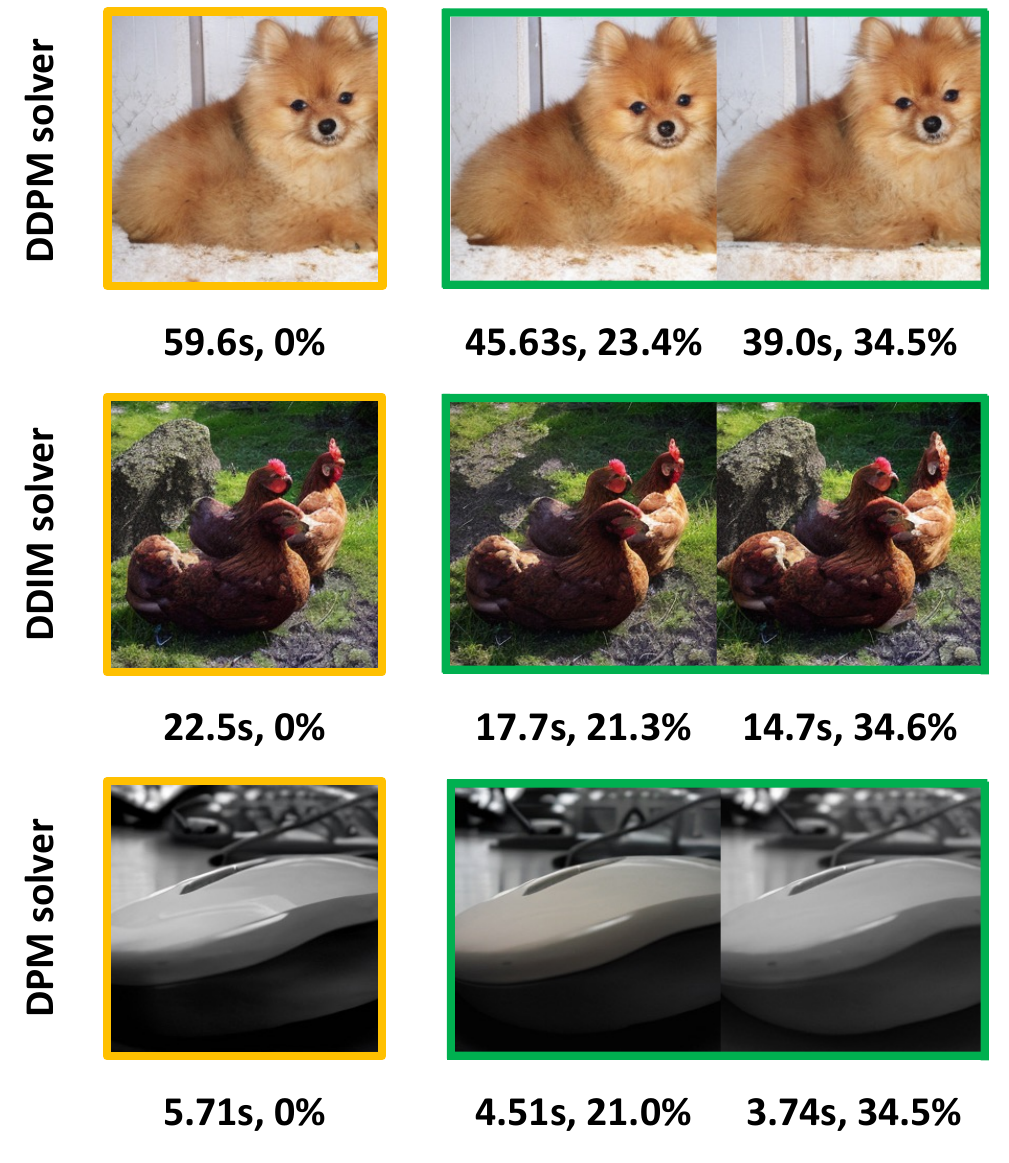}\vspace{1em}
\caption{Images sampled from ASE-enhanced DiT model with diverse dropping schedules.}
\label{fig:dit_dropping}
\end{figure*}

\begin{figure*}[!hb]
\centering
\includegraphics[width=0.45\linewidth]{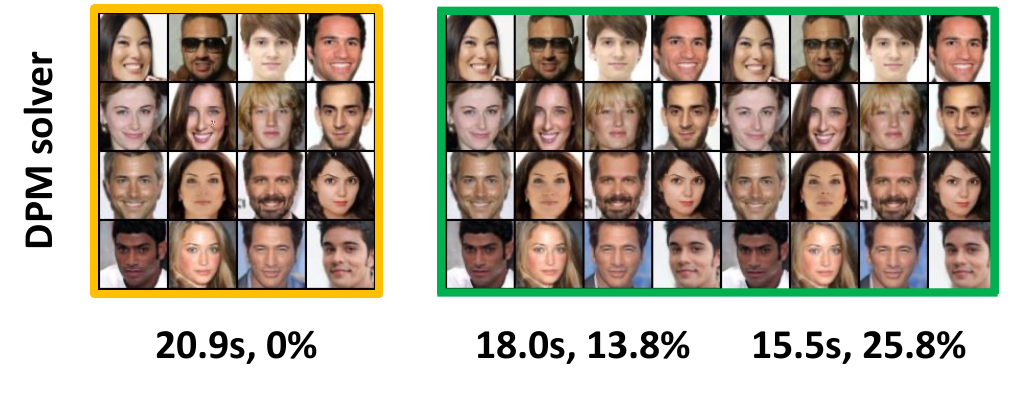}\vspace{1em}
\caption{Images sampled from ASE-enhanced U-ViT model with diverse dropping schedules.}
\label{fig:u_vit_dropping}
\end{figure*}

\begin{figure*}[!hb]
\centering
\includegraphics[width=0.45\linewidth]{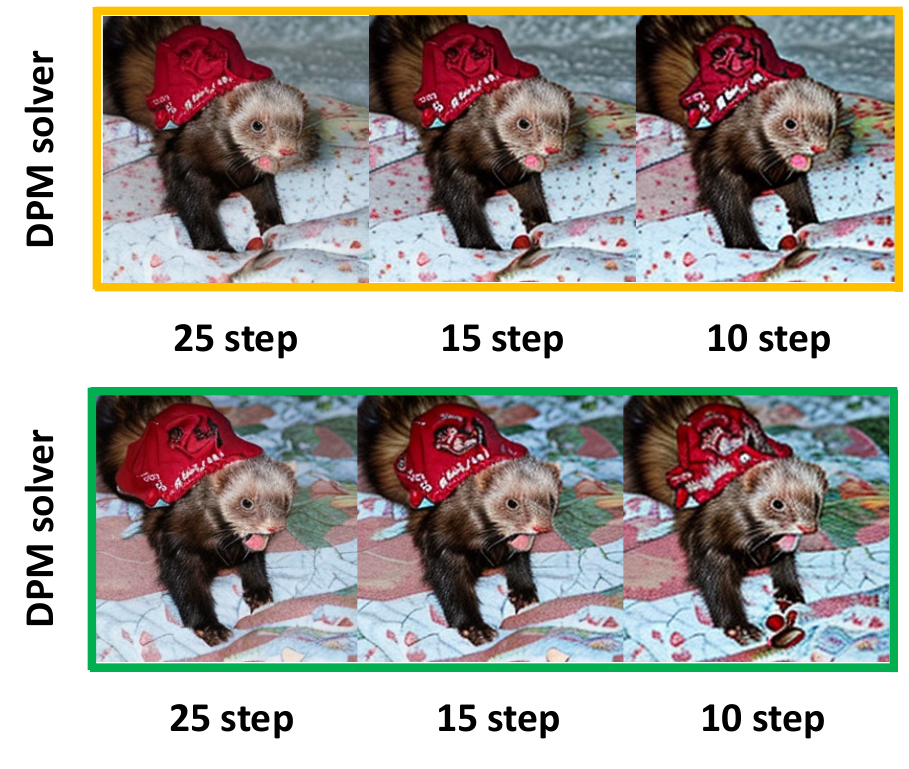}\vspace{1em}
\caption{Images sampled from the fine-tuned DiT model with DPM sampler.}
\label{fig:dit_step}
\end{figure*}

\begin{figure*}[!hb]
\centering
\includegraphics[width=0.45\linewidth]{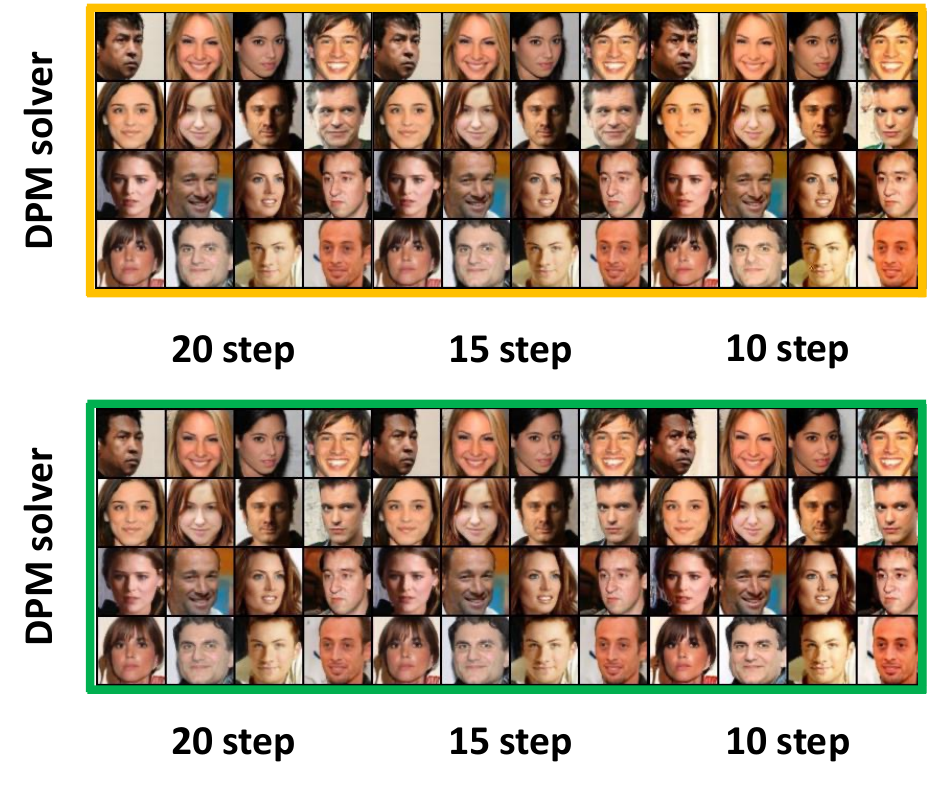}\vspace{1em}
\caption{Images sampled from the fine-tuned U-ViT model with DPM sampler.}
\label{fig:u_vit_step}
\end{figure*}

\begin{figure*}[t]
\centering
\includegraphics[width=0.5\linewidth]{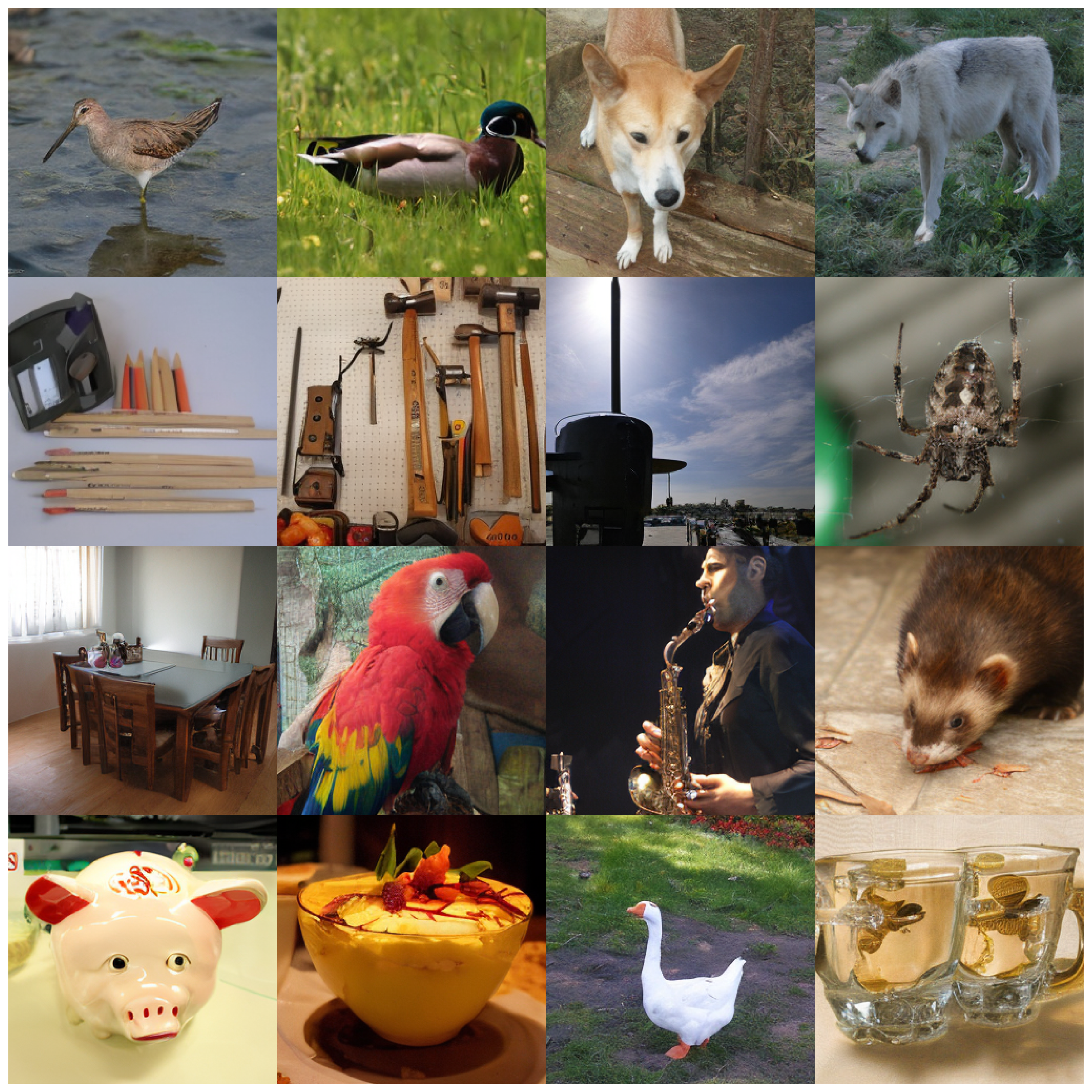}\vspace{1em}
\includegraphics[width=0.5\linewidth]{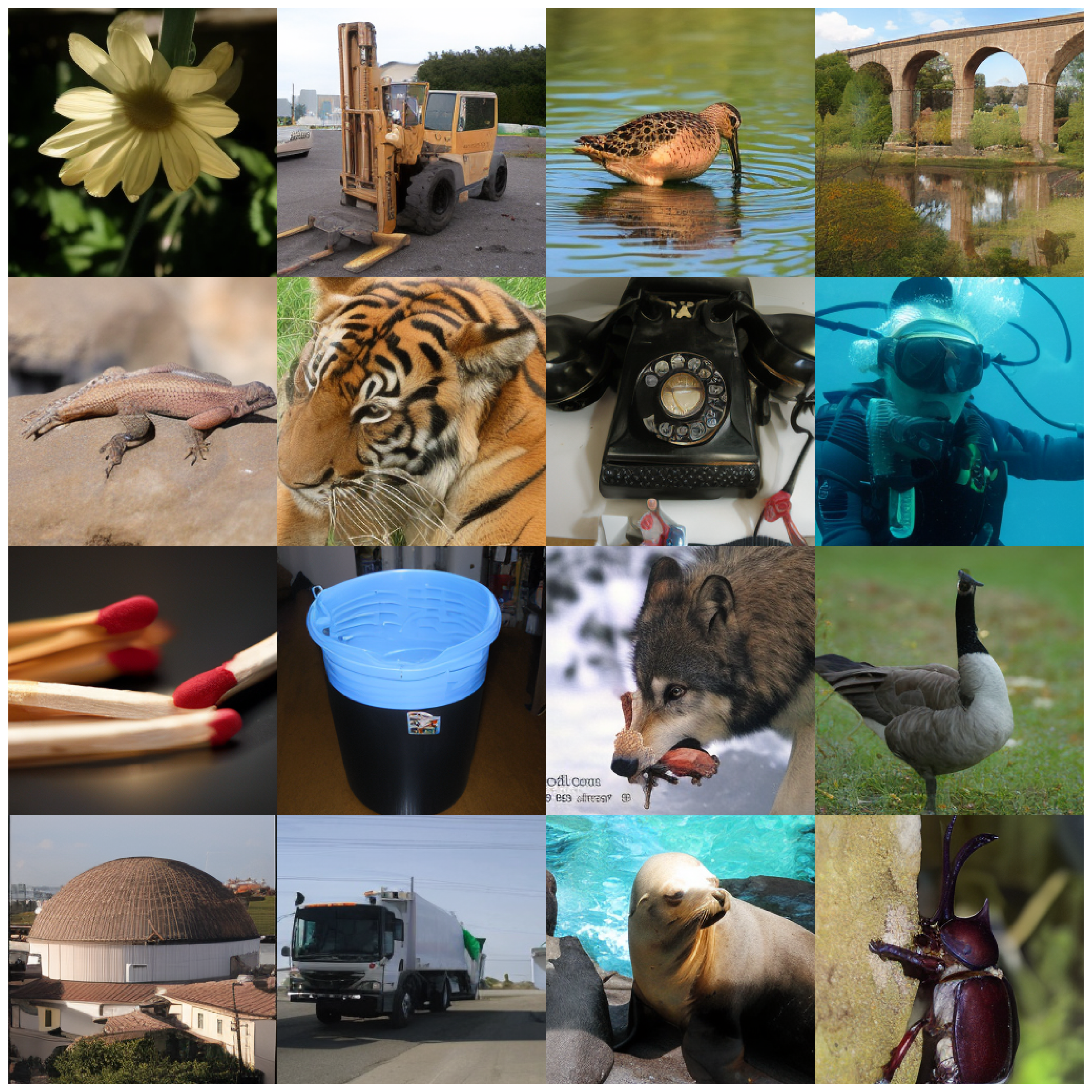}\vspace{1em}
\caption{Images sampled from the fine-tuned DiT model. Top: DDPM sampler-250 steps; Bottom: DDIM sampler-50 steps.}
\label{fig:dit_samples}
\end{figure*}

\begin{figure*}[t]
\centering
\includegraphics[width=0.5\linewidth]{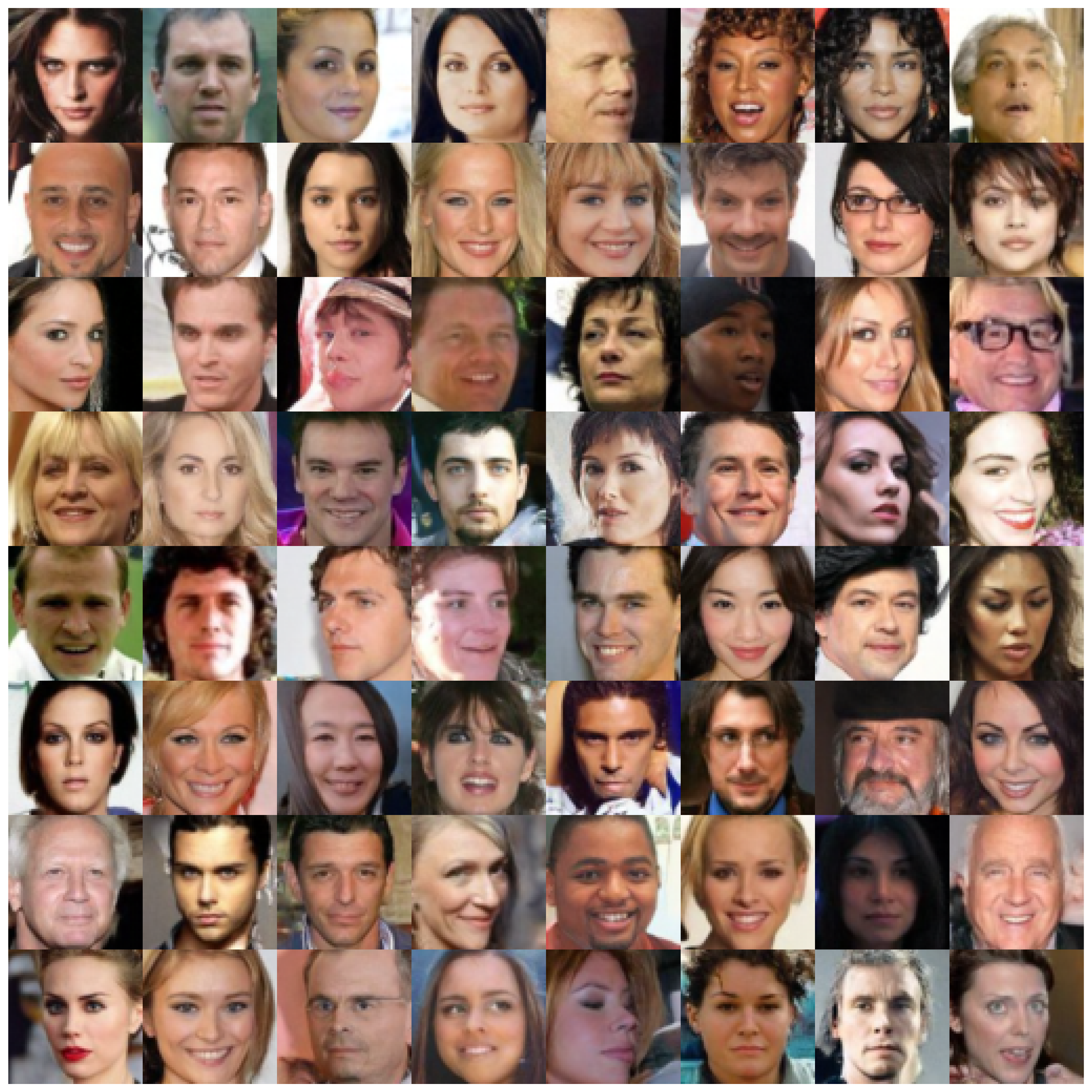}\vspace{1em}
\includegraphics[width=0.5\linewidth]{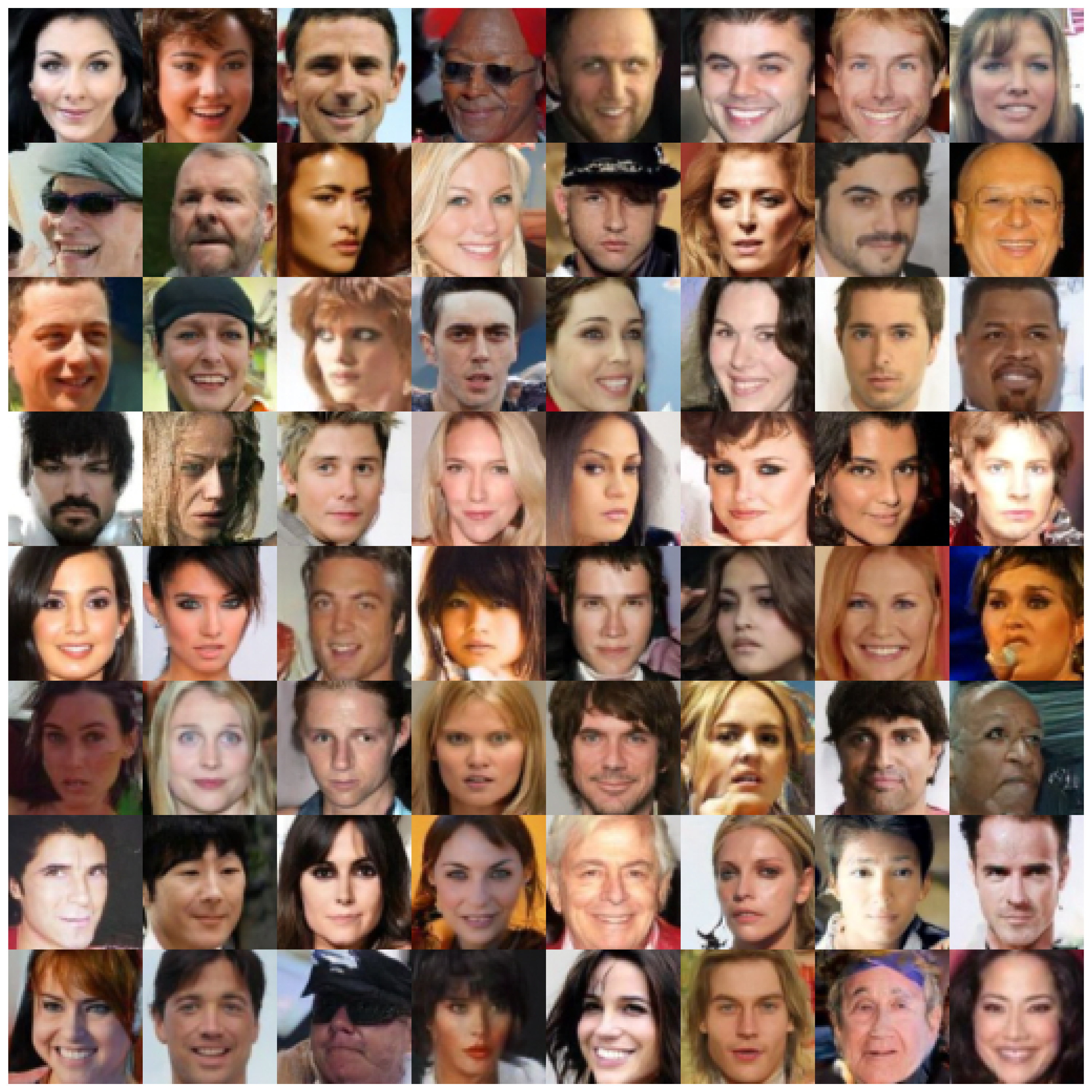}\vspace{1em}
\caption{Images sampled from the fine-tuned U-ViT model. Top: EM solver-1000 steps; Bottom: DPM solver-25 steps.}
\label{fig:u_vit_samples}
\end{figure*}

We present comprehensive experimental results, primarily including qualitative analyses. \cref{fig:dit_dropping} and \cref{fig:u_vit_dropping} shows the superior quality of generated samples under various dropping schedules. Additionally, in the \cref{fig:dit_step} and \cref{fig:u_vit_step}, we show the robustness of ASE across varing sampling timesteps. Notably, we provide visual representations of randomly generated images for each time-dependent early exiting schedule. In the \cref{fig:dit_samples}, it illustrates the results obtained by sampling from fine-tuned DiT checkpoint using both the DDPM and DDIM sampler. Similarly, in the \cref{fig:u_vit_samples}, it exhibits the results obtained by sampling from fine-tuned U-ViT checkpoint using both the EM and DPM sampler. 

\end{document}